\documentclass[letterpaper]{article} 
\usepackage{aaai24}  
\usepackage{times}  
\usepackage{helvet}  
\usepackage{courier}  
\usepackage[hyphens]{url}  
\usepackage{graphicx} 
\usepackage{natbib}  
\usepackage{caption} 
\usepackage{algorithm}
\usepackage{algorithmic}

\usepackage{newfloat}
\usepackage{listings}
\DeclareCaptionStyle{ruled}{labelfont=normalfont,labelsep=colon,strut=off} 
\floatstyle{ruled}
\newfloat{listing}{tb}{lst}{}
\floatname{listing}{Listing}
\usepackage{amsmath}
\usepackage{amsthm}
\usepackage{booktabs}
\usepackage{algorithm}
\usepackage{algorithmic}

\usepackage{makecell}
\usepackage{multirow}
\usepackage{amssymb}
\usepackage{latexsym}
\usepackage{mathrsfs}
\usepackage{xfrac}
\usepackage{float}
\usepackage{subfig}
\usepackage{color}

\newtheorem*{defn}{Definition}

\title{Knowledge-Aware Neuron Interpretation for Scene Classification}
\author {
    Yong Guan\textsuperscript{\rm 1,\rm 2},
    Freddy L\'{e}cu\'{e}\textsuperscript{\rm 3}$^*$,
    Jiaoyan Chen\textsuperscript{\rm 4},
    Ru Li\textsuperscript{\rm 2}$^*$,
    Jeff Z. Pan\textsuperscript{\rm 5}\thanks{Corresponding authors}
}
\affiliations {
    \textsuperscript{\rm 1} Department of Computer Science and Technology, Tsinghua University, China\\    
    \textsuperscript{\rm 2}School of Computer \& Information Technology, Shanxi University, China\\
    \textsuperscript{\rm 3}Inria, France\\
    \textsuperscript{\rm 4}Department of Computer Science, The University of Manchester, UK\\
    \textsuperscript{\rm 5}School of Informatics, The University of Edinburgh, UK \\
    gy2022@mail.tsinghua.edu.cn, freddy.lecue@inria.fr, liru@sxu.edu.cn, j.z.pan@ed.ac.uk,
}

\usepackage{bibentry}

\begin{document}

\maketitle

\begin{abstract}
Although neural models have achieved remarkable performance, they still encounter doubts due to the intransparency. To this end, model prediction explanation is attracting more and more attentions. However, current methods rarely incorporate external knowledge and still suffer from three limitations: (1) \textbf{Neglecting concept completeness}. Merely selecting concepts may not sufficient for prediction. (2) \textbf{Lacking concept fusion}. Failure to merge semantically-equivalent concepts. (3) \textbf{Difficult in manipulating model behavior}. Lack of verification for explanation on original model. To address these issues, we propose a novel knowledge-aware neuron interpretation framework to explain model predictions for image scene classification. Specifically, for concept completeness, we present core concepts of a scene based on knowledge graph, ConceptNet, to gauge the completeness of concepts. Our method, incorporating complete concepts, effectively provides better prediction explanations compared to baselines. Furthermore, for concept fusion, we introduce a knowledge graph-based method known as Concept Filtering, which produces over 23\% point gain on neuron behaviors for neuron interpretation. At last, we propose Model Manipulation, which aims to study whether the core concepts based on ConceptNet could be employed to manipulate model behavior. The results show that core concepts can effectively improve the performance of original model by over 26\%.
\end{abstract}


\section{Introduction}
\label{sec:intro}

\begin{figure*}[htbp]
  \centering
  \includegraphics[width=0.85\textwidth]{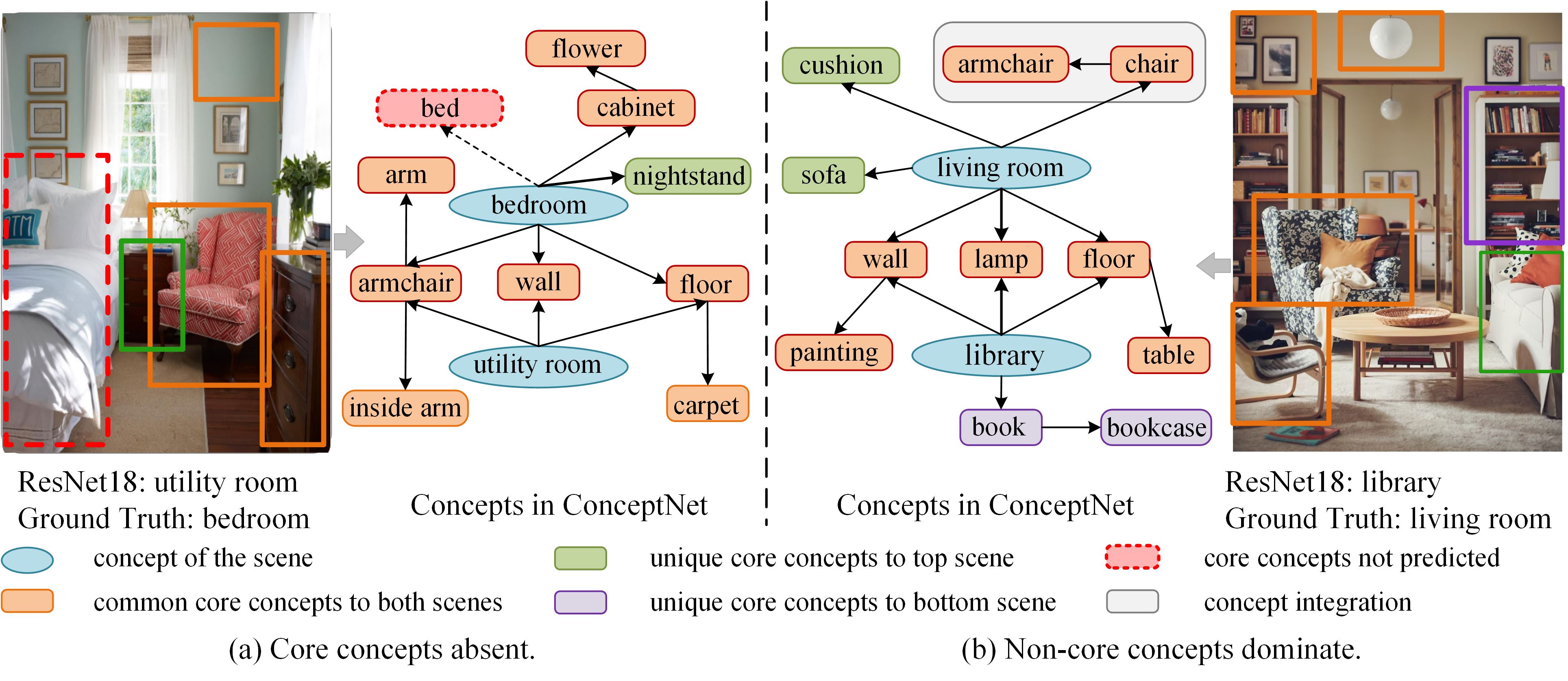}\\
  \caption{Example of false prediction explanations\protect\footnotemark. Concepts that are essential to the meaning of a scene are called core concepts, such as \textit{bed} in scene \texttt{bedroom}. Concepts that are not necessary for understanding the scene and can be omitted or ignored are called non-core concepts, such as \textit{book} and \textit{bookcase} in \texttt{living room}. } 
  \label{example_display}
\end{figure*}

Deep neural network (DNN) architectures are designed to be increasingly sophisticated, with scaled up the model size, and have been achieving unprecedented advancements in various areas of artificial intelligence~\cite{thoppilan2022lamda,openai2023gpt4}. Despite their strengths, DNNs are not fully transparent and often perceived as ``black-box'' algorithms, which can impair users' trust and hence diminish usability of such systems \cite{brik2023survey}. As shown in Figure \ref{example_display}(a), the model predicts the image as \texttt{utility room}, which is different from the ground truth (target label) \texttt{bedroom}. However, it is unclear why the model predicts this label, making it hard to understand, debug and improve.

There has been a growing interest in exploring explanations of model predictions~\cite{CLPHC2018,DZZC+2019}, which, generally speaking, 
could be categorized into two methods: functional analysis and decision analysis~\cite{abs-2102-01792}. 
Functional analysis methods try to capture overall behavior by investigating the relations between the decision and the image, using saliency map~\cite{Akhtar_Jalwana_2023}, occlusion techniques~\cite{kortylewski2021compositional}, and rationale~\cite{jiang-etal-2021-alignment}. 
Such methods typically lack in-depth understanding of internal modules of the model, 
often failing to provide comprehensive 
insights into the decision-making process. 
The decision analysis methods explore explanation by analyzing the internal components' behavior, such as decomposing the network classification decision into contributions of input elements~\cite{MONTAVON2017211,tian2020mane}. 
Furthermore, studying neuron-level explanation enables more accurate orientation and editing of the decision-making process \cite{Teotia_2022}. However, they do not offer the most intuitive explanations that are easily understandable to humans, and the link between the decision and internal components is not obvious.

Some studies attempt to utilize concepts to enhance the interpretation of model decisions, establishing relations between the decision and the input image through a selected number of concepts, such as ACE~\cite{ghorbani2019towards}, ConceptSHAP~\cite{yeh2020completeness}, and VRX~\cite{nrc_2021}. Although the decision is explained by presenting a set of concepts found within image, these methods still exhibit certain three key limitations.
(1) \textbf{Neglecting concept completeness}. These methods select a set of concepts salient to the corresponding scene, but they do not guarantee that these concepts are sufficient to explain the prediction. As shown in Figure \ref{example_display}(a), the model selects a set of salient concepts, including \textit{armchair}, \textit{floor}, \textit{wall}, and more. However, their prediction mislabels scene as \texttt{utility room} instead of \texttt{bedroom}, due to an incomplete concept set~\cite{ZGPZ*2015} that overlooks the \textit{bed} concept.
(2) \textbf{Lacking concept fusion}. These methods merely group segments based on resemblance, but they do not merge semantically-equivalent concepts. As depicted in Figure \ref{example_display}(b), concepts \textit{armchair} and \textit{chair} can be fused~\cite{WFZPR2015}, as  they convey identical meanings.
(3) \textbf{Hard to manipulate model behavior}. These methods mainly focus on explanation, but they do not provide guidance on how to rectify mistakes made by the original model. 

\footnotetext{ (a) The model does not predict the CC \textit{bed} of \texttt{bedroom}, and thus mistakenly predicts the scene to be \texttt{utility room}. (b) The model mainly focuses on non-core concepts, including \textit{book} and \textit{bookcase}, which are CC of scene \texttt{library}.}

To address the above problems, we propose a novel knowledge-aware neuron interpretation framework for scene classification.
As well-defined knowledge will help to further enhance the model explanation and facilitate human understanding. 
Specifically, for \textbf{concept completeness}, we present core concepts (CC) of a scene to gauge the completeness of concepts. CC refers to the fundamental elements that collectively constitute the scene~\cite{kozorog2013towards}; e.g., \textit{bed}, \textit{cabinet}, \textit{armchair}, \textit{floor}, \textit{wall}, \textit{nightstand} and \textit{lamp} are the CC of scene \texttt{bedroom}. 
To formulate the CC, we leverage 
knowledge graphs (KG)~\cite{PVGW2017,Pan2017b}, such as ConceptNet. 
Additionally, we introduce the MinMax-based NetDissect method to establish links between neuron behavior and concepts at the neuron level.
As for \textbf{concept fusion}, we introduce a Concept Filtering method, which effectively merge semantically-equivalent concepts based on ConceptNet in order to enhance the existing neuron interpretation.
Furthermore, for \textbf{manipulate model behavior}, we propose Model Manipulation, which aims to study whether the CC obtained from ConceptNet could be employed to manipulate model behavior, such as identifying the positive/negative neurons in model, integrating CC into original model design phase. 
At last, our method, integrating CC, will help to answer a variety of interpretability questions across different datasets, such as ADE20k and Opensurfaces, and multiple models, such as ResNet, DenseNet, AlexNet and MobileNet.

\begin{itemize}
    \item Do complete concepts base on core concepts provide benefit to model prediction explanation? We propose core concepts of scene derived from ConceptNet, along with three evaluation metrics, to establish the link between decisions and concepts in image. The experimental results show that our method, integrating complete concepts, achieves better results than the existing methods.

    \item Furthermore, do external knowledge through concept fusion improve existing neuron interpretation? We propose \textit{Concept Filtering} method, which produces over 23\% point gain on neuron behaviors for neuron interpretation. 

    \item In addition, do explanations based on core concepts contribute to model performance? We propose both unsupervised and supervised methods based on core concepts extracted from ConceptNet to manipulate model behavior. The overall results prove that core concepts and related explanation metrics can help optimise the original model, leading to 26.7\% of performance improvement\footnote{Code and data are available at: https://github.com/neuroninter\\pretation/EIIC}.

\end{itemize}

\section{Preliminaries}

\begin{figure*}[t]
\centering
    \begin{minipage}[t]{0.2\linewidth}
        \centering
        \includegraphics[width=\textwidth]{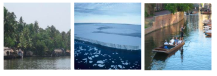}
        \centerline{(a) Input Image $x$.\label{fig:iou:a}}
    \end{minipage}%
    \begin{minipage}[t]{0.2\linewidth}
        \centering
        \includegraphics[width=\textwidth]{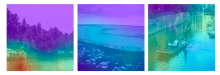}
        \centerline{(b) Activation $F_{483}(x)$.\label{fig:iou:b}}
    \end{minipage}%
    \begin{minipage}[t]{0.2\linewidth}
        \centering
        \includegraphics[width=\textwidth]{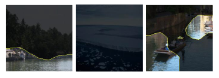}
        \centerline{(c) Activation over $T_{483}$\label{fig:iou:c}}
    \end{minipage}%
    \begin{minipage}[t]{0.2\linewidth}
        \centering
        \includegraphics[width=\textwidth]{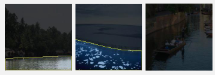}
        \centerline{(d) Concept \texttt{water}.\label{fig:iou:d}}
    \end{minipage}%
    \caption{{Unit $483$ of layer $4$ in ResNet-18, as a \texttt{water} detector (d) with a IoU score of $.14$ for pixelwise annotated input images $x$ in (a) wrt. the upscaled unit activation map (b), determined by areas of significative activation (c).}}
    \label{fig:iou}
\end{figure*}

\subsection{Neuron Interpretation}
Neuron interpretation aims to improve the interpretability of models by understanding the neuron behavior. Observations of neurons (a.k.a hidden units) in neural networks have revealed that human-interpretable concepts sometimes emerge as individual latent variables. Thus, a pioneer work on interpreting neurons~\cite{Bau2017NetworkDQ} designed a network dissection (\textit{NetDissect}) tool to quantify the interpretability of a model and particularly its neurons. 

Given a neural network $f$  trained and used for prediction, $f$ maps an image $x_{i}$ to a latent representation that is also known as neuron features or units (e.g., unit $483$ of ResNet-18 layer $4$ in Figure \ref{fig:iou}), denoted as $\{ f_{1}, f_{2}, ..., f_{n}\}$, where $n$ denotes the dimension and $f_{t}$ ($1 \leq t \leq n$) is known as \emph{t}-th neuron features.
Given $C$ the  set of concepts of a given dataset, $L :(x_{i},c) \mapsto \{0,1\}$ is a  concept function which indicates whether an image (region) $x_{i}$ is an instance of a concept $c \in C$; e.g., 
$L(x_{i},water)=1$ means $x_{i}$ is an image (region) containing water. \textit{NetDissect} computes the most relevant concepts  from $C$ to the neuron feature $f_t$ over the set of images $x$:   
\begin{equation}\small
    \mathit{NWD} (f_{t}, x, C) = \mathit{argmax} \{ \sigma (A_{t}(x_i), L(x_i,c))\}
\end{equation}
where $\sigma$ is a measure function, such as intersection over union (IoU) and Jaccard index, while $A_{t}(x_i)$ is the activation of $f_{t}$ for $x_i$, and can be scaled up to the mask resolution using bilinear interpolation. The unit $f_{t}$ is regarded as a detector which selects the highest scoring concept. For example, unit $483$ in Figure \ref{fig:iou} is regarded as a \texttt{water} descriptor (d) for images (a) wrt. activation (b) using threshold (c).

\subsection{Problem Statement}
Let $\mathcal{D}=\{ x_{1}, x_{2}, ..., x_{|\mathcal{D}|}\}$ be a set of images, $C$ be the overall concept set, and $Y=\{y_{1}, y_{2}, ..., y_{|Y|}\}$ be a set of scenes.
Each image $x_{i} \in \mathit{\mathcal{D}}$ belongs to a scene $y_{j} \in Y$ and contains multiple concepts representing the scene in $x_{i}$; 
e.g., the image of Figure \ref{example_display}(a) is labelled as scene \texttt{bedroom} and contains concepts such as \emph{wall}, \emph{lamp}, \emph{armchair}. 
For each scene $y_{j}$, it has multiple images in $\mathcal{D}$, denoted as $\mathcal{D}_{y_{j}} \subseteq \mathcal{D}$.
$C_{y_{j}}\subseteq C$ refers to the set of associated concepts of $y_{j}$. 
$LC(x_{i})$ is the set of learning concepts of neuron features $f_{t}$ by computing the correlation between $f_{t}$ and the corresponding concept set. 
In the rest of paper, assumed that the KG contains all concepts in $C$.\footnote{In practice, if some concepts in $C$ are not in the KG, we could align them to similar concepts in the KG.}

For each image $x_{i}$, it has a label $y_p$ predicted by $f$ and a target (ground truth) label $y_j$.
In this paper, we consider three tasks: \textit{(T1)}  
model prediction explanation: explaining why $f$ predicts $x_{i}$ as $y_p$, and why the prediction is correct (i.e., $y_p = y_j$) or wrong (i.e., $y_p \neq y_j$); \textit{(T2)} neuron interpretation: studying the effectiveness of external knowledge used in \textit{T1} on existing neuron interpretation; 
\textit{(T3)} model manipulation: exploiting CC used in \textit{T1 and T2} to optimise model performance.

\section{Approach}

In this section, we present the knowledge-aware framework to address the three tasks mentioned above.
Task \textit{T1}, corresponding to the limitation of concept completeness, contains three sections: MinMax-based NetDissect, Core Concepts, and Model Prediction Explanations. Task \textit{T2}, corresponding to the limitation of concept fusion, is detailed in Concept Filtering. Task \textit{T3}, corresponding to the limitation of manipulate model behavior, is detailed in Model Manipulation.

\subsection{MinMax-based NetDissect}
\label{neuronexp}
MinMax-based NetDissect aims to learn which concepts are closest to the neuron. 
Following existing work~\cite{Bau2017NetworkDQ,guan-etal-2023-trigger}, we use NetDissect to evaluate the alignment between each hidden unit and a set of concepts.
Note that  the original NetDissect method computes the general neuron behavior on the dataset level while ignoring features that are unique and useful for an individual image prediction. 
In contrast, we aim to learn the neuron behavior for individual scene, e.g., \texttt{bedroom}, to see whether concepts in scene can help explain model prediction.
To achieve this, we propose a new variant MinMax-based NetDissect method to learn the neuron behavior for individual image.
Formally, given $f$   a neural network, $f_{t}$ the \emph{t}-th neuron in intermediate layer, 
$C_{y_{j}}$   the associated concepts of $y_{j}$,    $x_{i}$ the target image, $\sigma$ measure function, and $A_{t}$ the activated neuron features  of the \emph{t}-th neuron, $L(x_i,c)$ the concept function where $c$ is a concept in $C_{y_j}$,  we have: 
\begin{equation}\label{minmax_netdissect}\small
  \mathit{MM\text{-}NWD} \ (f_{t}, x_i, C_{y_j}) = \mathit{Ths} \{  \sigma (A_{t}(x_{i}),  L(x_i,c) )\}
\end{equation}
 We use IoU as the measure function $\sigma$.
Thus, the concepts $LC(x_{i})$ learned by a target neuron can be obtained from concept selection strategy\footnote{In NetDissect, they directly use the concept with highest score of neuron. However, neurons do not express a single concept, but make predictions from multiple concepts \cite{comexpneu2020}.}$\mathit{Ths\{\cdot\}}$.
We consider three ways of selecting concepts that neuron learns.

(1) \emph{Whole layer}: all concepts with IoU scores larger than 0 are regarded as valid concepts; 

(2) \emph{Highest IoU}: only select the concept with the highest IoU score, treating the neuron as a concept detector. 

(3) \emph{Threshold}: only utilize the concepts with IoU scores higher than  a MinMax-based threshold that we compute as follows: (a) select the concept with the highest IoU for each neuron; (b) use  the lowest IoU value among the IoU values of the selected concepts as the threshold.

\subsection{Core Concepts}
\label{core_concepts_definition}

Core concepts (CC) are the fundamental elements that are strong candidates for   scene concepts introduced in~\cite{kozorog2013towards}; e.g.,    \textit{bed}, \textit{cabinet}, \textit{armchair}, \textit{floor}, \textit{wall}, \textit{nightstand} and \textit{lamp} are the CC of scene \texttt{bedroom}. A well-defined KG, such as ConceptNet, contains essential concepts in the scene and facilitates human understanding. Therefore, we leverage ConceptNet to help to define two types of CC to address the challenge of concept completeness, including scoping core concepts (SCC) and identifier core concepts (ICC). 
Informally speaking, 
the SCC for a scene involves concepts that are shared between the scene-related concepts within the dataset and the scene-related concepts within the KG.
On the other hand, the ICC for a scene involves concepts that are uniquely suited to a particular scene, such as \textit{bed}, \textit{nightstand} for \texttt{bedroom}.
Between SCC and ICC, 
SCC takes into account concept coverage for the scene, while ICC considers the concepts specific to the scene.
Next, two types of CC are 
defined as follows.

\begin{defn}{\bf (Scoping Core Concepts)}
Given a scene $y_j\in Y$  ($j\in\{1,...,|Y|\}$), its associated concepts in whole dataset $\mathcal{D}$ are denoted as $C_{y_{j}}$, and $RC(y_{j},\mathcal G)$ is the concepts from KG $\mathcal G$ that are related to $y_{j}$. We define the {\bf scoping core concepts} for scene $y_j$ as follows: $SCC(y_{j},\mathcal G)=RC(y_{j},\mathcal G)\cap C_{y_j}$.
\end{defn}

\begin{defn}{\bf (Identifier Core Concepts)}
Given a scene $y_j\in Y$  ($j\in\{1,...,|Y|\}$), its associated images and concepts in whole dataset $\mathcal{D}$ are $\mathcal{D}_{y_{j}}$ and $C_{y_{j}}$, respectively. 

\begin{itemize}
  \item 
  Concepts of a scene that obtained form dataset. 
  $Count(y_j,p)\subseteq C_{y_{j}}$ is the set of overlapping ground truth concepts, from $C_{y_{j}}$, over at least $p\%$ of the images in  $\mathcal{D}_{y_{j}}$,
  \item 
  Specificity concepts for a scene obtained form dataset.
  $P_c$ is the highest percentage such that, for any $i,j$, $i \neq j$, $Count(y_i,P_c) \neq Count(y_j,P_c)$.
  \item 
  Concepts of a scene obtained form dataset and KG.
  $SCount(y_j,\mathcal G, p)$ is the set of overlapping ground truth concepts, from $(RC(y_j,\mathcal G) \cap C_{y_j}) \cup TopkOfCount(y_j)$,  over at least  $p\%$ of the images in  $\mathcal{D}_{y_{j}}$, where $TopkOfCount(y_j)$ is the set of the top $k$ concepts of $Count(y_j$, $P_c)$ and $\mathcal{G}$ is Knowledge Graph,
  \item 
   Specificity concepts of a scene obtained form dataset and KG.
  $P_{sc}$ is the highest percentage such that, for any $i,j$, $i \neq j$, $SCount(y_i,\mathcal G, P_{sc}) \neq SCount(y_j,\mathcal G,P_{sc}) )$.
\end{itemize}
We define the {\bf identifier core concepts} for  scene $y_j$ as follows:  $ICC(y_j,\mathcal G)=SCount(y_j,\mathcal G, P_{sc})$.
\end{defn}

We consider the balance between concepts in the KG and annotated concepts  of $y_j$, by including the top $k$ (in our experiments, $k=2$) most popular concepts, no matter whether they are in the KG or not. As ICC is more specific, it often has a smaller size than SCC.

\subsection{Model Prediction Explanations}
\label{model_pred_exp}

Model Prediction Explanations aims to utilize 
CC to establish the link between decisions and concepts in image. For this purpose, we propose the following metrics accordingly.

\textbf{Prediction explanations (PE)} are explanations provided together with predictions, with ground truth (target) scene unknown.
Given an image $x_{i}$, concepts $LC(x_{i})$ learned by neurons, scene $y_j$, and its core concepts $CC_l(y_j)$, where $CC_l \in \{\mathit{SCC}, \mathit{ICC} \}$. We propose the consistency metric (similarity metric, difference metric) for measuring the consistency (similarity, difference, resp.):  
\begin{equation}\label{metric_cm}\small
  CM(x_{i},y_j) = \frac{|LC(x_{i}) \cap CC_l(y_j)|}{|CC_l(y_j)|}
\end{equation}

\begin{equation}\label{metric_sm}\small
  SM(x_{i},y_j) = \frac{|LC(x_{i}) \cap CC_l(y_j)|}{|LC(x_{i}) \cup CC_l(y_j)|}
\end{equation}

\begin{equation}\label{metric_dm}\small
  DM(x_{i},y_j) = \frac{|LC(x_{i}) \setminus CC_l(y_j)|}{|CC_l(y_j)|}
\end{equation}

Note $y_j$ is the predicted scene. The larger (smaller) the $CM$ and $SM$ ($DM$) scores become, the smaller the gap between the learned concepts and scene.

\textbf{Post-prediction explanations (PPE)} are explanations when both predicted and target scene are known.
Given an image $x_{i}$ of scene $y_t$, and the scene $y_p$ predicted by model, the first task here is to explain why the prediction is wrong, i.e., why $y_t\neq y_p$. 
One would expect that the LC should be closer to the predicted scene (i.e.,  $CM(x_{i},y_p) > CM(x_{i},y_t)$ and $SM(x_{i},y_p) > SM(x_{i},y_t)$) and be more different from the target scene (i.e., $DM(x_{i},y_p) > DM(x_{i},y_t)$).   
The consistency metric for set $D_f$ of false predictions (as scene $y_p$) $CM^{FP}$ can be defined as follows:
\begin{equation}\label{false_mm}\small
  CM^{FP} = \frac{|\{x_{i} \in D_f | CM(x_{i},y_p) > CM(x_{i},y_t) \}|}{|D_f|}
\end{equation}
The difference and similarity
metrics, denoted as $DM^{FP}$ and $SM^{FP}$ can be defined respectively.

Given an image $x_{i}$ of scene $y_t$, 
the second task here is to explain why the prediction is correct. We propose to   compare the set of images with true prediction ($D_t$) against those with false prediction ($D_f$), 
with the expectation that the consistency metric over the correctly predicted images $CM^{TP}$ of $D_t$ should be larger than that over the falsely predicted images $CM^{T\_FP}$: $CM^{TP} > CM^{T\_FP}$.\footnote{The symbol $T$ refers to calculating true prediction explanation.} Thus, the consistency metric for the set $D_t$ of correctly  predicted images and that for the set $D_f$ of falsely predicted images in scene $y_t$ can be defined as follows :
\begin{equation}\label{}\small
  CM^{TP} = \frac{\Sigma_{x_{i}\in D_t} CM(x_{i},y_t)}{|D_t|}
\end{equation}
\begin{equation}\label{}\small
CM^{T\_FP} = \frac{\Sigma_{x_{i}\in D_f} CM(x_{i},y_t)}{|D_f|}
\end{equation}
Similarly, we can define similarity metric for $D_t$ and $D_f$, denoted as $SM^{TP}$ and $SM^{T\_FP}$, respectively, with the expectation 
that $SM^{TP}>SM^{T\_FP}$.

\subsection{Neuron Interpretation via Concept Filtering}
\label{sec:filtering}
This section aims to optimize neuron interpretation by merging semantically-equivalent concepts based on ConceptNet. In context of image classification and object detection, there could be a large number of concepts and many of which might have similar semantics, e.g. \emph{armchair} and \emph{chair}. This could lead to misleading or even wrong explanations for predictions. To address this challenge, given each set of scene associated concepts $C_{y_j}$, we compute the embeddings of the   concepts in $C_{y_j}$ and align them to concepts in a KG like ConceptNet, using classic KG embeddings techniques, such as TransE,  Dismult and TransD, then group them w.r.t. their distances, into clusters $Cl_1(C_{y_j}), ..., Cl_r(C_{y_j})$. One can transform $C_{y_j}$ into $CF(C_{y_j})$ by selecting one representative concept in each cluster $Cl_i(C_{y_j}) (1\leq i \leq r)$  to represent all concepts in $Cl_i(C_{y_j})$. Our \emph{hypothesis} is that replacing $C_{y_j}$ with $CF(C_{y_j})$ could help optimise model prediction explanation and existing neuron interpretation.

\begin{table*}[t]
\small
\centering
\begin{tabular}{llccccccc}
\toprule
\multirow{2}*{LC}&\multirow{2}*{CC}&\multicolumn{3}{c}{False Prediction Explanation (\%)}&\multicolumn{4}{c}{True Prediction Explanation (\%)}\\
\cmidrule{3-5} \cmidrule(lr){6-9}
 & &$ \ \mathit{CM^{FP}} \ $ & $ \ \mathit{DM^{FP}} \ $ & $ \ \mathit{SM^{FP}} \ $ &$\mathit{CM^{TP}}$ & $\mathit{CM^{T\_FP}}$ & $\mathit{SM^{TP}}$ & $\mathit{SM^{T\_FP}}$\\
\hline
ConceptSHAP&--&51.93&43.24&50.87&21.04&18.51&19.68&17.32 \\
\hline
\multirow{2}*{CLIP-Dissect}&SCC &53.31&86.57&33.45 &12.64&11.78&13.72&13.16\\
&ICC & 46.94& 67.06&38.94 &33.31&32.44&21.76&21.17 \\
\hline
\multirow{3}*{Whole Layer}&Top\_10&43.04&19.13&42.77&11.29&6.12&4.38&2.49\\
&SCC & \textbf{78.51} & 87.30 & 69.85&12.95 & 9.94 & 8.17 & 6.85 \\
&ICC & 51.43 & 69.32 & 29.58& \textbf{53.73} & \textbf{47.33} & 22.76 & 22.13 \\
\hline
\multirow{3}*{Highest IoU}&Top\_10&34.61&19.12&34.56&9.20&5.18&7.51&4.03\\
&SCC & 65.77 & 85.76 & 64.07& 6.52 & 5.04 & 5.91 & 4.64 \\
&ICC & 49.42 & 67.76 & 42.24& 26.32 & 21.86 & 21.27 & 18.11 \\
\hline
\multirow{3}*{Threshold}&Top\_10&42.52&18,69&42.42&11.17&6.07&7.48&4.12\\
&SCC & 78.03 & \textbf{87.38} & \textbf{72.13}& 11.83 & 9.22 & 10.09 & 8.07  \\
&ICC & 50.32 & 69.81 & 34.11 & 49.60 & 44.13 & \textbf{34.97} & \textbf{32.34}\\
\bottomrule
\end{tabular}
\caption{ Results of false and true prediction explanation. Top\_10 means the top 10 concepts of scene as CC.}
\label{table_false_true_exp}
\end{table*}

\subsection{Model Manipulation}
\label{sec:improvement}

Model Manipulation aims to study whether the CC could help to manipulate model behavior, including Neuron Identifying via CC and Re-training via CC. 
In addition, we propose using PE metrics for re-training.

\textbf{Neuron Identification via CC}
aims to identify the positive and negative neurons by calculating contribution score to see the model behavior.
The contribution score for the neurons $f_t$, over images $x_{i}$ in $y_j$ with true prediction, can be calculated as follows:
\begin{equation}\label{neu_contri}\small
  \mathit{Con\_Score(f_t)} = \sum_{x_{i} \in D_{y_{j}}^{T}} (P(x_{i}, CC_{l}) - N(x_{i}, CC_{l}))
\end{equation}
where $P(x_{i}, CC_{l})$ and $N(x_{i}, CC_{l})$ are the number of LC in and not in $CC_{l}$, respectively. 

For true prediction, we disable top-k positive neurons (by setting the neuron features to 0 \cite{comexpneu2020}) for the scene and see whether the model still correctly predicts the scene.
For false prediction, we disable top-k (in our evaluation, $k = 20$) negative neurons for the scene and see whether the model can make better prediction.

\textbf{Re-training via CC}
aims to integrate CC into original model design phase to further improve its performance. 
In the original models, the training objective is scene loss $\mathit{\mathcal{L}_{s}}$. We add another core concept loss:  
\begin{equation}\small
    \mathit{\mathcal{L}_{c}} = - \sum \log \mathcal{P}(c^{*}|\theta)
\end{equation}
where $c^{*} \in C$ is the golden concept. 
For example, given scene \texttt{bedroom} with concepts \textit{bed, armchair and fridge}, the new overall objective will let model pay more attention to the CC, such as \textit{bed} and \textit{armchair}.

\textbf{Re-training via PE} 
aims to utilize the explanation metrics as features to optimise the original model. We use a classical classifier SVM \cite{svm_2006}, but not an arbitrary neural network, as it will not introduce unexplained factors. 
For training the classifier, we utilize three types of features: (1) the features of metrics CM, SM and DM; (2) the MRR (mean reciprocal rank) feature which integrates the three metrics over all scenes; (3) the hidden states which learned by the original model.

\section{Experiments}\label{sec:ex}

\subsection{Datasets}
For testing, we use two scene datasets ADE20k \cite{ADE20kdataset} and Opensurfaces \cite{bell14intrinsic}.  
ADE20k is a challenging scene parsing benchmark with pixel-level annotations, which contains 22,210 images.
There are 1,105 unique concepts in ADE20k, categorized by scene, object, part, and color, and each image belongs to a scene. We utilize the version from existing work CEN \cite{comexpneu2020}. Opensurfaces is a large database created from real-world consumer photographs with pixel-level annotations. It contains 25,329 images which are annotated with surface properties, including material, color and scene.

\subsection{Do Completing Concepts Based on Core Concepts Provide Benefit to Model Prediction Explanation?}
\label{sec:mpe}

\textit{\textbf{Yes. The overall results show that our method, integrating complete concepts, achieves better results than existing methods across false and true prediction}}. In details, we first report the results of false/true prediction explanation, 
and then enhance the model prediction explanation by integrating concept filtering. Furthermore, we conduct experiments on different models. 

\textbf{Results of False Prediction Explanation}
For false prediction explanation, we expect to have higher scores on the three metrics ($CM$, $DM$, $SM$). The higher the score, which means the better the explanations. 
The results are reported in Table \ref{table_false_true_exp}, and we have the following three observations.

(1)
When compared to the results of the baseline method Top\_10, both SCC and ICC achieve significant better results, indicating our proposed method is effective and reasonable.

(2)
All the best scores for false prediction explanation (across $CM^{FP}$, $DM^{FP}$ and $SM^{FP}$) come from SCC. The reason is that SCC has broader coverage than ICC: if some concepts are not in SCC, then they are most likely to be incorrect. On the other hand, ICC is more specific, thus it might exclude some (partially) correct concepts.

(3) Among the three methods to represent neurons' learned concepts in section MinMax-based NetDissect, the threshold-based method achieves better results, demonstrating that our method can explain 
false predictions   well. 

(4) The results from ConceptSHAP~\cite{yeh2020completeness} and CLIP-Dissect~\cite{oikarinen2023clipdissect} are in line with the trends in NetDissect, and all satisfy our assumptions.

\begin{table}[t]
\small
\centering
\begin{tabular}{lccc}
\toprule
\multicolumn{4}{c}{Neurons' concepts: whole layer (\%)} \\
\midrule
 Methods&$ \ \mathit{CM^{FP}} \ $ & $ \ \mathit{DM^{FP}} \ $ & $ \ \mathit{SM^{FP}} \ $ \\
\midrule
Top\_10&58.76&26.14&58.82\\
SCC & \textbf{81.17}&\textbf{85.97}&\textbf{74.74}\\
ICC & 55.73&69.85&37.39\\
\bottomrule
\end{tabular}
\caption{ Integrating concept filtering for false prediction. }
\label{combine_con_filter_false}
\end{table}

\begin{table}[t]
\small
\centering
\begin{tabular}{lcccc}
\toprule
\multicolumn{5}{c}{Neurons' concepts: whole layer (\%) } \\
\midrule
\multirow{2}*{Methods} & \multicolumn{2}{c}{Consistency Metrics} & \multicolumn{2}{c}{Similarity Metrics} \\
\multicolumn{1}{c}{}&$\mathit{CM^{TP}}$ & $\mathit{CM^{T\_FP}}$ & $\mathit{SM^{TP}}$ & $\mathit{SM^{T\_FP}}$ \\
\midrule
Top\_10&15.79&9.88&6.23&3.59\\
SCC & 25.31&19.69&18.23&15.01\\
ICC & \textbf{66.60} &\textbf{59.78} &\textbf{41.19} & \textbf{38.16}\\
\bottomrule
\end{tabular}
\caption{ Integrating concept filtering for true prediction. }
\label{combine_con_filter_true}
\end{table}

\textbf{Results of True Prediction Explanation}
For true prediction explanation, we expect to observe that $\mathit{CM^{TP}}$ and $\mathit{SM^{TP}}$ are larger than $\mathit{CM^{T\_FP}}$  and $\mathit{SM^{T\_FP}}$ respectively. The bigger the scores as well as the gap between $\mathit{CM^{TP}}$ and $\mathit{CM^{T\_FP}}$ and between $\mathit{SM^{TP}}$ and $\mathit{SM^{T\_FP}}$, the better the results are.
As a whole, the results in Table \ref{table_false_true_exp}, 
ICC achieves the better results. Although the gap between $\mathit{SM^{TP}}$ and $\mathit{SM^{T\_FP}}$ for top 10 (Highest IoU) is bigger, these $\mathit{SM^{TP}}$ and $\mathit{SM^{T\_FP}}$ scores are very low.

\textbf{Integrating Concept Filtering for Model Prediction Explanation}
Tables \ref{combine_con_filter_false} and \ref{combine_con_filter_true} show the results for false prediction explanation and true prediction explanation when using concept filtering to simplify the concept sets.
For false prediction, SCC achieves the best performance compared to ICC and Top\_10. For true prediction, once again the results of $\mathit{CM^{TP}}$ and $\mathit{SM^{TP}}$ are larger than $\mathit{CM^{T\_FP}}$ and $\mathit{SM^{T\_FP}}$ respectively. In addition, results are better than that without concept filtering in Table \ref{table_false_true_exp}. 

\begin{table}[t]
\small
\centering
\begin{tabular}{lcccc}
\toprule
Models&CC &$ \ \mathit{CM^{FP}} \ $ & $ \ \mathit{DM^{FP}} \ $ & $ \ \mathit{SM^{FP}} \ $ \\
\midrule
\multirow{2}*{ResNet-50}&SCC & \textbf{80.85}&\textbf{89.36}&\textbf{74.46} \\
&ICC & 53.19&69.15&36.17 \\
\midrule
\multirow{2}*{DenseNet-161}&SCC & \textbf{78.63}&\textbf{81.32}&\textbf{45.65} \\
&ICC & 55.47&57.54&21.29 \\
\midrule
\multirow{2}*{AlexNet}&SCC & \textbf{76.58} &\textbf{83.26} & \textbf{71.34}\\
&ICC & 60.23 &68.33  & 31.57 \\
\bottomrule
\end{tabular} 
\caption{Results of false prediction explanation on different models and utilize the MinMax-based threshold to learn the neurons' concepts.}
\label{mpx_diff_model_false}
\end{table}

\textbf{Model Prediction Explanation on Different Models}
We further implement our method on different architectures to verify the generalization. We randomly select 1000 samples from the ADE20k data for the experiment by considering the effect of time efficiency.

The results of false prediction explanation are shown in Table \ref{mpx_diff_model_false}.
From the CC perspective, 
SCC has better results than ICC over every model, which is similar to the observation over ResNet-18 from
Table \ref{table_false_true_exp}.
The results of SCC on ResNet-50 achieve the best performance across all models.

\begin{table}
\small
\centering
\begin{tabular}{ccccc}
\toprule
Cluster Nb. & TransD & Dismult & ProjE & TransE \\
\midrule
160 & +17.01 & +22.20 & +19.94 & +20.11 \\
165 & +18.11 & +22.84 & +17.48 & +25.77 \\
167 & +18.44 & +23.42 & +18.03 & +26.05 \\
168 & +21.78 & +23.15 & +17.80 & \textbf{+26.31} \\
169 & +20.72 & +22.84 & \textbf{+23.86} & +26.01 \\
170 & +20.90 & +23.22 & +22.35 & +25.36 \\
175 & +21.27 & \textbf{+23.88} & +21.74 & +22.47 \\
180 & \textbf{+25.29} & +23.07 & +23.04 & +22.63 \\
185 & +24.10 & +23.34 & +22.49 & +22.32 \\
\bottomrule
\end{tabular}
\caption{IoU gain (\%) of different clusters.}
\label{graphEmbeddings}
\end{table}

\subsection{Do External Knowledge and Concept Fusion Improve Existing Neuron Interpretation?}
\label{sec:comp}

\textit{\textbf{Yes. Our method, Concept Filtering, merges semantically-equivalent concepts based on ConceptNet, which effectively produces over 23\% point gain on neuron behaviors for neuron interpretation.} }
As KG embedding techniques could have an impact on the number of optimal clusters, as well as on the interpretability of neurons, we ran some experiments with ResNet-18 over the ADE20k dataset to evaluate their impact. In particular we evaluated the impact of TransE \cite{bordes2013translating},  Dismult \cite{yang2014embedding}, ProjE \cite{shi2017proje} and TransD \cite{ji2015knowledge} on the (1) optimal number of clusters, and (2) quality of interpretability, measured using IoU similarly described as in CEN. The final number of cluster also captures the final number of core concepts to be considered for explanation, as a cluster is described by a unique concept in ConceptNet.
The knowledge graph used for computing the embeddings is a subset of ConceptNet. In particular, we extracted all concepts in ADE20k, as well as direct 1-hop and 2-hop neighbors of ADE20k concepts in ConceptNet. We applied fuzzy matching for $0.1\%$ of ADE20k concepts due to some misalignment between  concepts in ADE20k and ConceptNet.

\begin{figure*}[t]
\centering
\begin{tabular}{cc}
\begin{minipage}[t]{0.32\linewidth}
   \includegraphics[width = 1\linewidth]{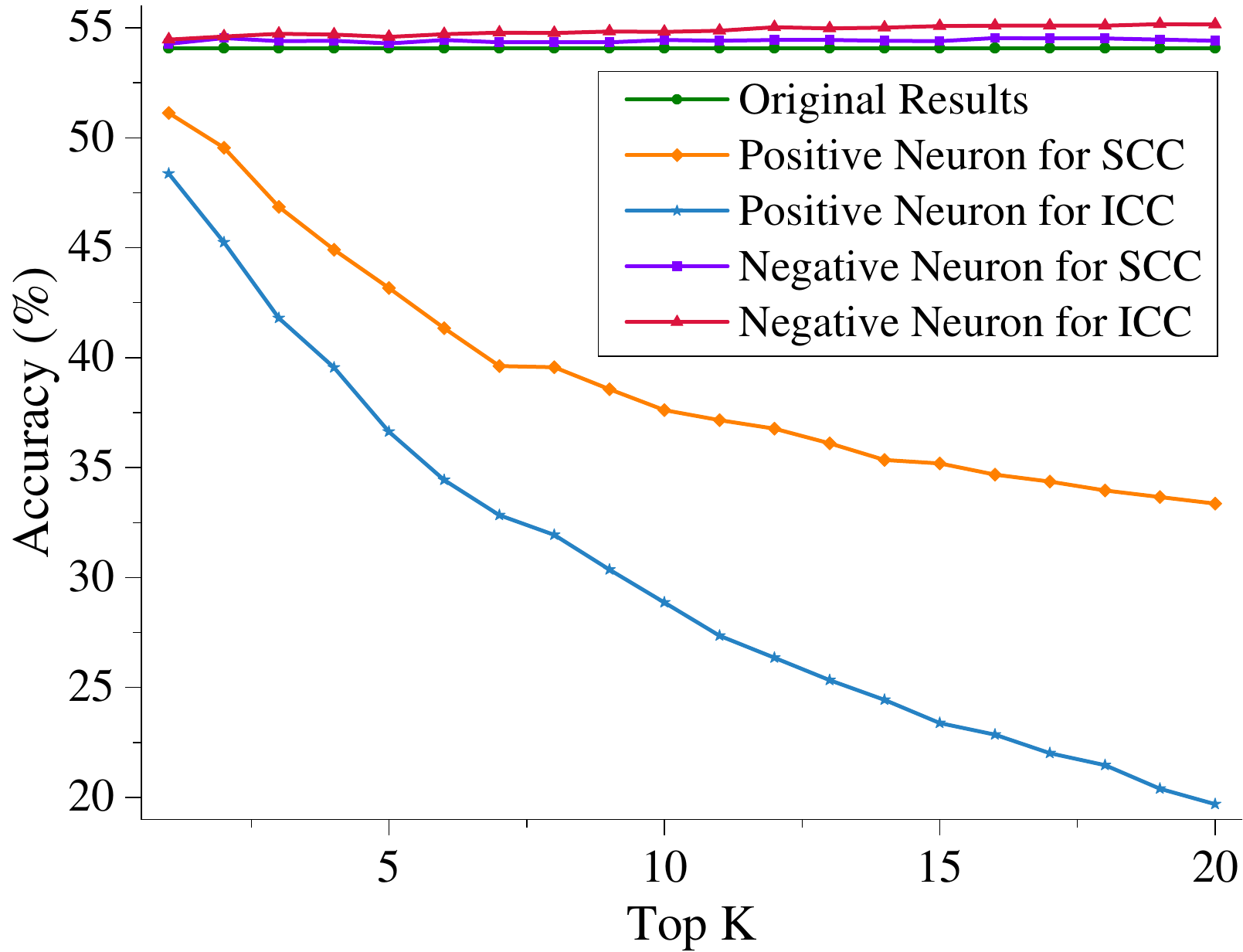}
   \caption{Results on ADE20k.}
   \label{neu_eff_ADE20k}
\end{minipage}
\hfill
\begin{minipage}[t]{0.32\linewidth}
   \includegraphics[width = 1\linewidth]{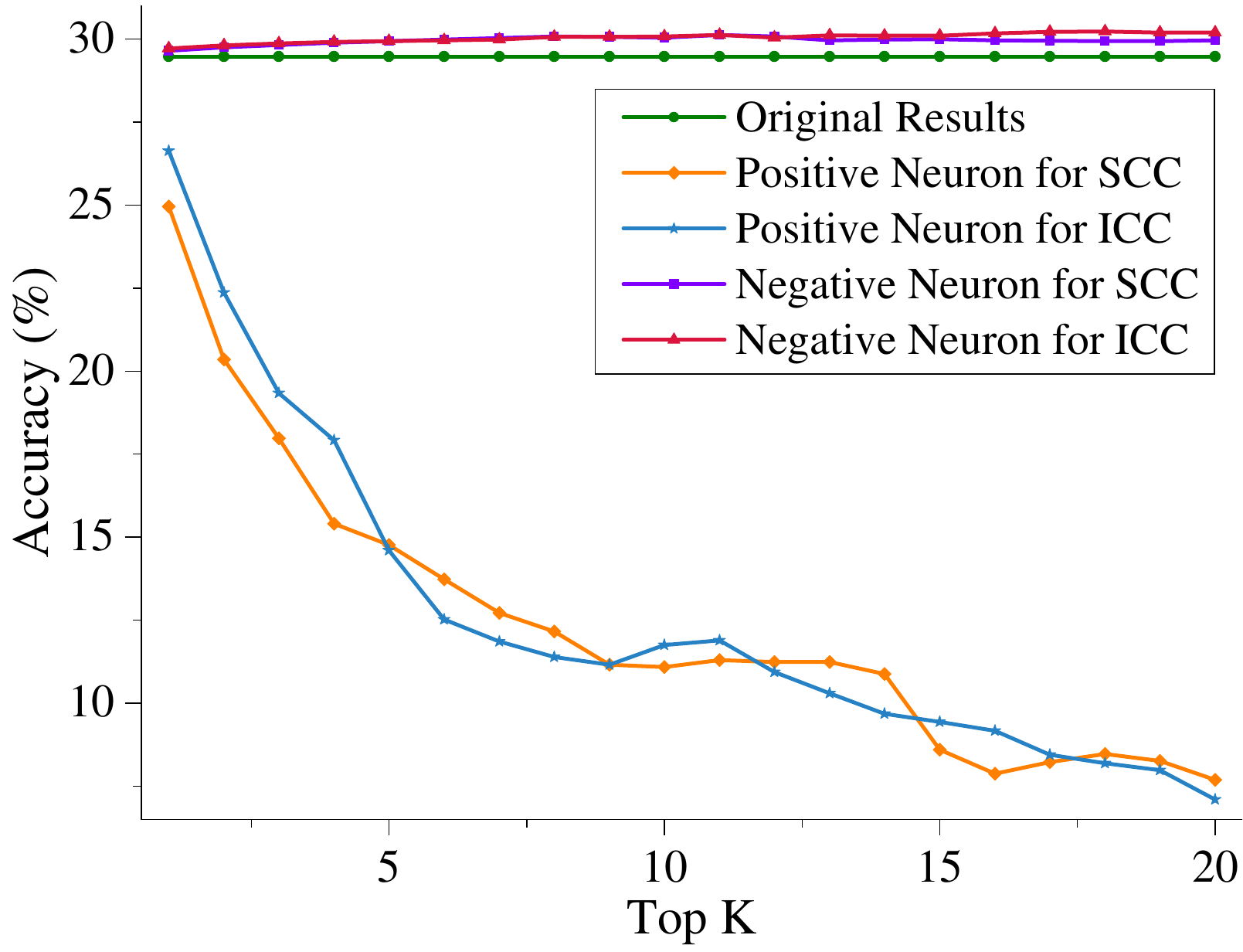}
   \caption{Results on Opensurface.}
   \label{neu_eff_opensurface}
\end{minipage}
\hfill
\begin{minipage}[t]{0.32\linewidth}
   \includegraphics[width = 1\linewidth]{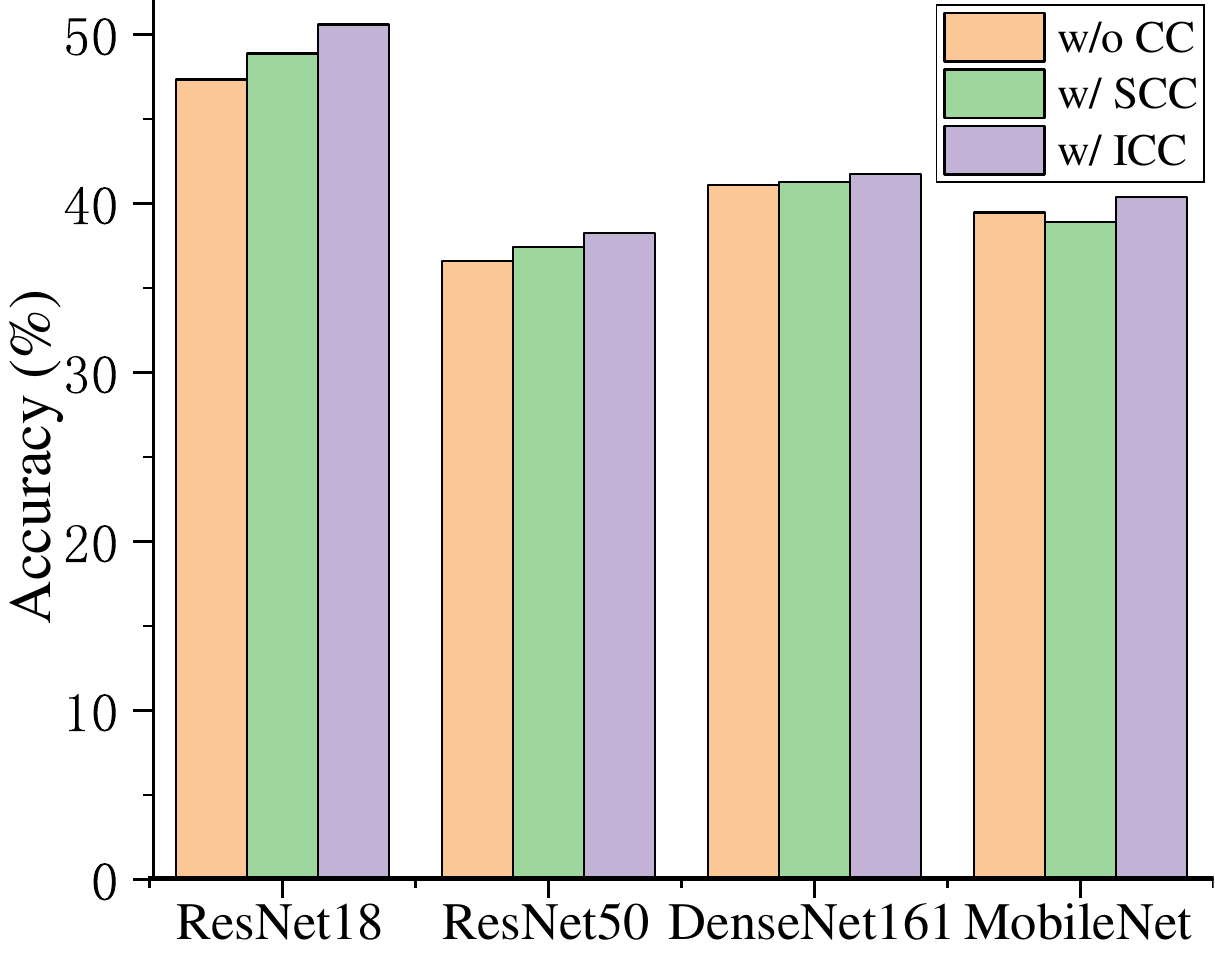}
   \caption{Model re-training via CC.}
   \label{retrain_model}
\end{minipage}
\end{tabular}
\end{figure*}

The IoU gain is measured by capturing the interpretability improvement from (A) concept with no clustering strategy to (B) concept with a $k$-clustering strategy with ($k$: Cluster Nb.) using Embeddings. The IoU gain is defined as $\sfrac{(B - A)}{A}$.
Table \ref{graphEmbeddings} captures the main results. 
We can see that:
(1) Fusing of semantically-equivalent concepts leads to significant performance enhancement across all methods. In essence, reducing cluster number exposes more interpretable units in the neural model.
(2) Among various embedding techniques, TransE outperforms others, achieving a remarkable 26.3\% improvement with 168 clusters compared to the non-clustering strategy, i.e., $512$ neurons in the context of ResNet-18.
(3) From a clustering perspective, optimal performance arises within the 168-180 class range. Fewer than 168 clusters imply fusion of less semantically related concepts, while over 180 clusters suggest the disregard of partially semantically-equivalent concepts.


\subsection{Do Explanations Based on Core Concepts Contribute to Model Performance?}
\label{eva_model_per}

\textit{\textbf{Yes. We show that core concepts and related explanation metrics can help optimise the model, leading to 26.7\% of performance improvement.}} After verifying the effectiveness of CC on explanations, a natural question is whether CC can be further used to help manipulate model behavior.

\textbf{Results of Neuron Identification via CC}
Figure \ref{neu_eff_ADE20k} shows the model performance when we disable the positive neurons or negative neurons. We have the following three observations: (1) when negative (resp. positive) neurons are disabled, the model performance is improved (resp. decreases),
proving that the CC facilitates identifying important neurons during decision-making of the model;
(2) model accuracy tends to decrease as the number of inhibitory
positive neurons increases;
(3) compared to SCC, ICC can better identify both the positive and negative neurons.
As shown in Figure \ref{neu_eff_opensurface}, we can also see
that the model performance decreases when we disable the  positive neurons.
Note that the accuracy on ADE20k decreases more, indicating that more positive neurons could be detected on ADE20k.

When inhibiting negative neurons, the model performance is improved on both ADE20k and Opensurfaces.
The results on Opensurfaces do not show as much growth as the results on ADE20k.
This is  probably because Opensurfaces mainly focuses on annotating the surface property, such as material, which makes the core concept of different scenes with limited differentiation. For example, concepts \textit{painted}, \textit{wood} will be core concepts for most scenes, such as \texttt{living room}, \texttt{family room}, \texttt{office} and \texttt{staircase}.
From the overall experimental results, with the help of core concepts, our method can effectively identify the positive and negative neurons, and then augment the model performance.

\textbf{Results of Re-training via CC}
From Figures \ref{neu_eff_ADE20k} and \ref{neu_eff_opensurface}, we can see that the performance change of disabling negative neurons is not as large as disabling positive neurons on both datasets.
This is reasonable,
since the model we explain is trained with its parameters fixed and it is difficult to correct false predictions by only removing some negative neurons.
However, the improvements on different datasets still indicate that our method is effective to retrieve negative neurons.

To address this challenge, we re-train the initial models with the help of CC, and the results are shown in Figure \ref{retrain_model}. In Figure \ref{neu_eff_ADE20k} and \ref{neu_eff_opensurface}, the experiments are based on the model ResNet18, and the results have improved about 1.3\% by removing the negative neurons. However, in Figure \ref{retrain_model}, the corresponding performance of ResNet18 has improved 3.27\%.
On the other models, the results by adding ICC are all improved.
Compared to SCC, utilizing the ICC for re-training model is more effective.

The above two parts mainly focus on manipulating the model behavior, such as identifying the positive/negative neurons and re-training from scratch.
In the following part, we further verify the effectiveness of explanation metrics CM, SM and DM; cf., Eq.~(\ref{metric_cm}), ~(\ref{metric_sm}) and ~(\ref{metric_dm}).

\begin{table}[t]
\centering
\begin{tabular}{lcc}
\toprule
 \multirow{2}*{Method}&\multicolumn{2}{c}{Accuracy (\%)} \\
 &ADE20k&Opensurfaces \\
\midrule
ResNet18&52.96&29.26\\
SVM (SCC)&66.54&31.62\\
SVM (ICC)&\textbf{67.11}&\textbf{32.27}\\
\bottomrule
\end{tabular}
\caption{Results of PE.} 
\label{results_wpe}
\end{table}
\textbf{Results of Re-training via PE}
The results are shown in Table \ref{results_wpe}, and ResNet18 is the fundamental model. The results are improved for SCC and ICC on both datasets.
ICC-based SVM on ADE20k achieves the best performance with 67.11, and outperforms the basic ResNet18 by a large margin of 14.15, which amouts to 26.7\% improvement.

\section{Conclusion }

In this study, we investigated knowledge-aware neuron interpretation for image scenes classification. To address the concept completeness, 
we proposed two types of core concepts (i.e., SCC and ICC) based on KGs. We show that 
SCC is effective on explaining false predictions, while ICC excels in neuron identification and model optimisation with concept loss. Our results also show that concept fusion and CC based metrics are effective for neuron interpretation and model optimisation, respectively, significantly outperforming state of the art approaches by over 20\%.



\section*{Acknowledgments}

We would like to thank the anonymous reviewers for their constructive comments and suggestions. This work is supported by a grant from the Institute for Tsinghua University Initiative Scientific Research Program, by the Science and Technology Cooperation and Exchange Special Program of Shanxi Province (No.202204041101016), by the Institute for Guo Qiang, Tsinghua University (2019GQB0003), by the Chang Jiang Scholars Program (J2019032), and the EPSRC project ConCur (EP/V050869/1).

\bibliography{aaai24}

\end{document}